\def\year{2018}\relax
\documentclass[letterpaper]{article} 
\usepackage{aaai18}  
\usepackage{times}  
\usepackage{helvet}  
\usepackage{courier}  
\usepackage{url}  
\usepackage{graphicx}  
\usepackage{soul}
\usepackage{natbib}
\usepackage[toc,acronym,nomain]{glossaries}
\usepackage{dialogue}
\newacronym{lgssm}{LG-SSM}{Linear Gaussian State Space Machine}
\newacronym{rnn}{RNN}{Recurrent Neural Network}
\newacronym{lstm}{LSTM}{Long-Short Term Memory}
\newacronym{ffnn}{FFNN}{Feed Forward Neural Network}
\newacronym{crf}{CRF}{Conditional Random Field}
\newacronym{hmm}{HMM}{Hidden Markov Model}
\newacronym{gmm}{GMM}{Gaussian Mixture Model}
\newacronym{memm}{MEMM}{Maximum Entropy Markov Model}
\newacronym{rcc}{RCC}{Region Connection Calculus}
\newacronym{rl}{RL}{Reinforcement Learning}
\newacronym{svm}{SVM}{support vector machine}
\newacronym{darpa}{DARPA}{Defense Advanced Research Projects Agency}
\newacronym{ecat}{ECAT}{Event capture and annotation tool}
\newacronym{gui}{GUI}{Graphical user interface}
\newacronym{sdk}{SDK}{Software development kit}
\newacronym{api}{API}{Application programming interface}
\newacronym{cwc}{CwC}{Communication with Computer}
\newacronym{lfd}{LfD}{Learning from Demonstration}
\newacronym{pbd}{PbD}{Program by Demonstration}
\newacronym{lfi}{LfI}{Learning from Interaction}
\newacronym{ai}{AI}{Artificial intelligence}
\newacronym{hci}{HCI}{Human-computer interaction}
\newacronym{ann}{ANN}{Artificial Neural Network}
\newacronym{rbf}{RBF}{Radial-Basis Function Network}
\newacronym{gru}{GRU}{Gated recurrent unit}
\newacronym{vs}{VoxSim}{Voxicon Simulator}
\newacronym{es}{EpiSim}{Epistemic Simulator}
\newacronym{xai}{XAI}{Explainable Artificial intelligence}
\newacronym{qr}{QR}{Qualitative Reasoning}
\newacronym{qsr}{QSR}{Qualitative Spatial Reasoning}
\newacronym{qs}{QS}{Qualitative spatial}
\newacronym{dbn}{DBN}{Dynamic Bayesian Network}
\newacronym{cnn}{CNN}{Convolutional Neural Network}
\newacronym{pov}{POV}{Point of view}
\newacronym{fov}{FOV}{Field of view}
\newacronym{s2s}{Seq2Seq}{Sequence-to-Sequence}
\newacronym{mse}{MSE}{mean squared error}
\newacronym{eci}{ECI}{Elementary Composable Ideas}
\newacronym{rgbd}{RGB-D}{RGB-Depth}
\newacronym{vr}{VR}{Virtual reality}
\newacronym{lll}{LLL}{Life-long learning}
\newacronym{l3d}{L3D}{Life-long Learning from Demonstration}
\newacronym{dnn}{DNN}{Deep Neural Network}
\newacronym{nlp}{NLP}{Natural Language Processing}
\newacronym{nlu}{NLU}{Natural Language Understanding}
\newacronym{nlg}{NLG}{Natural Language Generation}

\begin{document}
%
\title{Multimodal Interactive Learning of Primitive Actions}
\author{Tuan Do, Nikhil Krishnaswamy, Kyeongmin Rim, and James Pustejovsky \\
Department of Computer Science\\ Brandeis University
\\
Waltham, MA 02453 USA \\
  {\tt \{tuandn,nkrishna,krim,jamesp\}@brandeis.edu}  
}

\maketitle
\begin{abstract}
We describe an ongoing project in learning to perform primitive actions from demonstrations using an interactive interface. 
In  our previous work, we have used demonstrations captured from humans performing actions as training samples for a neural network-based trajectory model of actions to be performed by a computational agent in novel setups. 
We found that our original framework had some limitations that we hope to overcome by incorporating communication between the human and the computational agent, using the interaction between them to fine-tune the model learned by the machine.
We propose a framework that uses multimodal human-computer interaction to teach action concepts to machines, making use of both live demonstration and communication through natural language, as two distinct teaching modalities, while requiring few training samples.
\end{abstract}

\section{Introduction}

This work takes a position on learning primitive actions or interpretations of low-level motion predicates by the \gls{lfd} approach.  \gls{lfd} can be traced back to the 1980s, in the form of automatic robot programming (e.g., \cite{lozano1983robot}). Early \gls{lfd} is typically referred to as \textit{teaching by showing} or \textit{guiding}, in which a robot's effectors could be moved to desired positions, and the robotic controller records its coordinates and rotations for later re-enactment. In this study, we instead focus on a methodology to teach action concepts to computational agents, allowing us to experiment with a proxy for the robot without concern for physically controlling the effectors. 

As discussed in \cite{chernova2014robot}, there are typically two sub-categories of actions that can be taught to robots: 1) high-level tasks that are hierarchical combinations of lower-level motion trajectories; and 2) low-level motion trajectories, the focus of this study, that can be taught by using a feature-matching method. We have experimented with offline learning motion trajectories from captured demonstrations. This method has some limitations, including requiring multiple samples as opposed to one-shot (or few-shot) learning, and being unable to accept corrections to generated examples beyond training on more data \citep{do2018spring}.

There is a wealth of prior research on hierarchical learning of complex tasks from simpler actions \citep{veeraraghavan2007learning, dubba2015learning, wu2015watch, alayrac2016unsupervised, fernando2017unsupervised}. \gls{hmm} and \gls{gmm} have been used extensively in previous work \citep{akgun2012trajectories, calinon2007teacher} to model the learning and the reenacting components. \cite{do2018thesis} proposed to use \gls{rl} directed by a \textit{shape} rewarding function learned from sample trajectories. In contrast, we are investigating \gls{lfd} methods to teach primitive concepts such as \textit{move A around B}, or \textit{lean A against B}, or \textit{build a row}, or \textit {build a stack from blocks on the table}, and we propose a method to learn these action concepts from demonstrations, supplemented by interaction with the agent to verify or correct some of the suppositions that the agent learns while building a demonstration-trained model.

Recently, \cite{Mohseni-Kabir2018} proposed a methodology to jointly learn primitive actions and high-level tasks from visual demonstrations, with the support of an interactive question-answering interface. In this framework, robots ask questions in order to group primitive actions together to create high-level actions. In a similar fashion, \cite{lindes2017grounding} teach a task to robots, such as \textit{discard an object}, by giving step-by-step instructions built on top of the simple actions \textit{move}, \textit{pick up}, \textit{put down}; \cite{maeda2017active} showcase a system wherein a robot makes active requests and decisions in the course of learning primitive actions incrementally; \cite{tellex2011understanding} use probabilistic graphical models to ground natural language commands to the situation.  We think these types of communicative frameworks can be extended to learning low-level actions.

We view  this direction of interactive learning as particularly promising, where symmetric communication between humans and robots can be used to complement \gls{lfd} as a modality for teaching (cf. \cite{thomaz2008teachable}).

\subsection{Related research}

Naturalistic communication between humans tends to be multimodal \citep{veinott1999video,narayana2018cooperating}.
Human speech is often supplemented by non-verbal communication (gestures, body language, demonstration/``acting"), while linguistic expressions provide both transparent and abstract information regarding the actions and events in the situation, much of which is not readily available from demonstrations. 
Dynamic event structure \citep{PustMosz:2011,pustejovsky2013dynamic} is one approach to language meaning that formally encodes events as programs in a dynamic logic with an operational semantics. These events very naturally map to the sub-steps undertaken during the course of demonstrating a new action (``grasp", ``pick up", ``move to location", etc.).
"Motion verbs can be divided into complementary \textit{manner-} or \textit{path-}oriented predicates and adjuncts \citep{Jackendoff:1983}.
Changes over time can be neatly encapsulated in durative verbs as well as in gestures or deictic referents denoting trajectory and direction.
This allows humans to express where an object should be or go either using linguistic descriptions or by directly indicating approximate paths and locations.

Computational agents typically lack the infrastructure required to learn new concepts solely through linguistic description, often due to an inability to fully capture the intricate semantics of natural language. 
Thus, instead of providing verbose instructions, we treat the agent as an active learner who interacts with the teacher to understand new concepts, as suggested in \cite{chernova2014robot}.

Research in cognition (cf. \cite{agam2008geometric}) has investigated how humans imitate trajectories of different shapes,  giving a strong indication that we tend to be able to better remember trajectories that follow a consistent pattern (\textit{curvature consistency}). In this paper, we hypothesize that human primitive action concepts exhibit relatively transparent conceptual consistencies. We hope to learn these consistencies directly from data represented as sequential features on a frame-by-frame basis from demonstrations. 

\gls{qs} representations have  proven useful in analogical reasoning, allowing machine learning algorithms to perform generalizations over smaller amounts of data than required for traditional quantitative representations \citep{mclure2015extending}.  This allows it to serve as a bias in the model, reflecting human real-world knowledge as the human interlocutor would be expected to have.  Libraries of qualitative relations often draw extensively from longstanding observations of human bias in psychological experiments on spatial processing \citep{stevens1978distortions, gentner1981studies, weld2013readings}.  Thus a machine learning model trained on data that contains bias from humans should reflect those same biases and the model should demonstrate judgments of those of human subjects, in spatial reasoning as in other domains \citep{caliskan2017semantics}. In a limited sense, biases can be defined as modes (in a statistical sense) of a non-uniform distribution in a descriptive space. For example, our bias toward the value of 0 (in a range from $0$ to $\infty$) allows us to qualitatively distinguish between when two objects are \textit{externally connected} (i.e., with 0 or near-0 distance between them) or \textit{disconnected}. 

To summarize, we exploit the ability to express and describe actions in multiple modalities, in order to explore how to impart human concepts and biases to AI models. We use the QS representations as the abstraction level to bridge demonstrative data, action concepts, and the action language used and expressed by computational agents. 

\section{A learning scenario}


In the following conversation, both a human user and an avatar can observe a table containing both a red block and a green block:

\small
\begin{dialogue}
\speak{Human} Can you \textbf{move the red block around the green block}?
\speak{Robot} Can you show me what ``around'' means?
\speak{Human} Yes. \direct{Slides the red block clockwise around the green block on the table}
\speak{Robot} Is the green block always stationary?
\speak{Human} Yes.
\speak{Robot} Is the red block always moving?
\speak{Human} Yes.
\speak{Robot} Is the red block always about the same distance from the green block?
\speak{Human} Yes.
\speak{Robot} Does the red block always move in the same direction relative to the green block?
\speak{Human} Yes.
\speak{Robot} OK.  Let me slide the red block around the green block.  \direct{Performs the action \textbf{slide the red block around the green block}, clockwise}
\end{dialogue}
\normalsize

\noindent
This scenario displays an effort to teach a machine a concept that is rather difficult to learn given only one or a few demonstrative samples. We also want to demonstrate the \textit{desiderata} of a machine learning system that can facilitate that learning:

\begin{itemize}
\item It can recognize pattern consistencies from feature data. Consistencies should be in formulaic representations that can be clearly articulated in natural language expressions. 
\item Pattern consistencies can be evaluated over multiple frames of the same demonstration. More importantly, a desirable framework should allow us to estimate the confidence of a pattern intended by the instructor.
\item The system should take a proactive role in interaction, by asking questions pertaining to patterns need to be verified. 
\item In terms of natural language interaction, the system has to be able to identify novel ideas as missing concepts in its semantic framework as well as to generate questions for verification of the recognized patterns. 
\end{itemize}

\section{Framework}

\begin{figure}[!h]
\centering
\vspace{-3mm}
\includegraphics[width=2.25in]{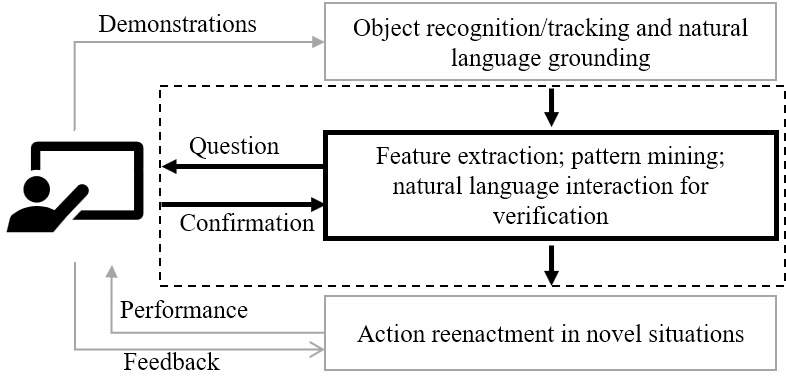}
\vspace{-3mm}
\caption{Interactive learning framework}
\label{fig:framework}
\end{figure}

Figure~\ref{fig:framework} depicts the architecture of our learning system.  For the top component, our experimental setup makes use of simple markers attached to objects for recognition and tracking. 
For natural language grounding, our proposal leverage the advancement of speech recognition \citep{Povey_ASRU2011} and syntactic analysis tools \citep{chen2014fast,reddy2017universal} to generate a grounded interpretation from spoken language. For the bottom component, we  discuss the use of ``mined patterns'' as constraints for action reenactment.  

The focus in this section is the middle component (inside the dotted box), including methods to mine pattern consistencies from demonstrative data, then pose generated natural language questions to human teachers to ask for confirmation of conceptual understanding, and use this understanding to constrain action performance when presented with a novel context or setup.

\subsection{Representations}

To represent pattern consistencies, we use a set of qualitative features that are widely used in the \gls{qsr} community. We have used these features as representation for action recognition \citep{do2017qsr} This is not intended to be an exhaustive set of features, and other feature sets, such as the \gls{rcc} \citep{cohn1997representing}, could be used as well. 

\small
\begin{itemize}
\itemsep-0.1em
    \item {\sc Cardinal direction} (CD) \citep{andrew1991qualitative}, transforms compass relations between two objects into canonical directions such as North, North-east, etc., producing 9 different values, including one for where two locations are identical. This feature can be used for the relative direction between two objects, for an object's orientation, or for its direction of movement.
    \item {\sc Moving} or {\sc static} (MV) measures whether a point is moving or not.
    \item {\sc Qualitative Distance Calculus} (QDC) discretizes the distance between two moving points, e.g., the distance between two centers of two blocks.
    \item {\sc Qualitative Trajectory Calculus} (Double Cross): $QTC_C$ is a representation of motions between two objects by considering them as two moving point objects (MPOs) \citep{delafontaine2011implementing}. We  consider two feature types of this set, whether two points are moving toward each other ($QTC_{C1}$) or whether they are moving clockwise or counterclockwise w.r.t. each other ($QTC_{C3}$). 
\end{itemize}
\normalsize

These qualitative features can be used to create the formulaic pattern consistencies that we are looking for in the previous discussion. All features can be interpreted as univariate or multivariate functions binding to tracked objects at a certain time frame. Hereafter, let $f^k_t(d)(x,y)$ denote the qualitative feature extracted from demonstration $d$ at frame $t$ of feature type $k$ between two objects $x$ and $y$. 

\subsection{Pattern mining}

The following describes some of the pattern consistencies that we are hoping  to learn from data:

\begin{itemize}
\item $f^k_0(d)(x,y) \, ? \, \alpha$ where $?$ can be any comparison operator $<, >, =, \leq, \geq, \neq$, and $\alpha$ is a constant value. This is a state to be satisfied at the start of a demonstration.
\item $f^k_{F}(d)(x,y) \, ? \, \alpha$ is a \textit{final} (F) state to be satisfied at the end of a demonstration.
\item ${\forall}t f^k_{t}(d)(x,y) \: ? \: \alpha$ describes a feature value that stays constant across all frames.
\item ${\forall}t f^k_{t}(d)(x,y) \: ? \: f^k_{t+1}(d)(x',y')$ describes a feature relationship between two consecutive frames. We allow a form of dynamic object binding so that it is not necessary that $(x,y) = (x',y')$, i.e., object binding is made by evaluating the demonstration $d$ at time $t$. However, in the example of ``Slide A around B'', (x,y) always bind to (A, B), because the system can map these directly from the instruction given to the demonstration.  
\item $f^k_{0}(d)(x,y) \: ? \: f^k_{F}(d)(x',y')$ relates features at the start (frame 0) and end (frame F) of the demonstration. 
\end{itemize}

These patterns $p \in \mathcal{P}$, where $\mathcal{P}$ is a partially ordered set. We define a \textit{precedence} relation, $ \preceq$ , so that two patterns can be compared. $p_1$ $ \preceq$  $p_2$ if $p_1$ is logically superseded by $p_2$. For example, $p_1 = f^k_0(d)(x,y) \, < \, \alpha$ takes precedence over $p_2 = f^k_0(d)(x,y) \, \neq \, \alpha$. 

To detect these patterns from data, we can define a function over patterns $q(p)$ that measures how confident we are that a pattern is intended in an action concept. 
This value should be higher when we have more demonstrations that exhibit the same pattern. 
Furthermore, $q$ should also give a pattern with a higher precedence a higher salience. 
The intuition is that if $p_1 \preceq p_2$, and we have $q(p_1) > t \wedge q(p_2) > t$ where $t$ is a confidence threshold, the system should ask for confirmation about $p_1$ before asking about $p_2$. 
When the teacher confirms $p_1$ to be true, the system then can take $p_2$ as trivially true.

Though attaining such a function is not trivial, we will give an illustrative example. Assume a 4-part quantization of QDC ("adjacent", "close", "far", "very far"), we define a bias $b$ over these values that characterizes the likelihood of a quantized region $v$ to be recognizable, for example $b=1/v$.
Finally, let $domain(p)$ be the range of the feature function $f$ that $p$ uses. 
Now, we define a heuristic function $q(p)$ as follows:
\begin{align*}
q(p) = \frac{probability(p) * bias(domain(p))}{|domain(p)|} 
\end{align*}
whereas $probability(p)$ is the probability that $p$ is correct among all samples, $bias(domain(p))=\sum_{v} b(v)$, and $|domain(p)|$ is the size of the domain. For example, if in 80\% of the samples, $f=0$ and in the remaining 20\%, $f=1$, we have $q(f=0)=0.8*1/1=0.8$, $q(f<=1)=1*(1+0.5)/2=0.75$, therefore, $q(f<=1)<q(f=0)$. If the ratio is 50:50, $q(f<=1)>q(f=0)$.

\subsection{Generating natural language questions }\label{ssec:nlg} 

Now that the system has patterns of qualitative features from observations, each associated with a \textit{confidence} score of consistency from the function $q$, it needs to confirm the intentionality of the patterns with the teacher.
To come up with a proper set of questions to ask, first the patterns need to be arranged into a queue in an order using the \textit{precedence} relation and confidence value, then the system forms natural language questions from the queue of patterns that need to be confirmed.
When the system gets a confirmation, it will iterate through the queue to remove now-implicit patterns. 

For instance, suppose $p_1$ ``\textit{object $X$ moving to the east}'' and $p_2$ ``\textit{$X$ moving in one direction}'' where $p_1 \preceq p_2$. When the teacher confirms $p_1$ first, the system does not have to ask for confirmation on $p_2$.
However, if the system is given only one demonstration, the function $q$ might not assign high confidence to $p_1 \equiv {\forall}t f^{MV\_dir}_{t}(d)(X) = EAST$ but higher value to $p_2 \equiv {\forall}t (f^{MV\_dir}_{t}(d)(X) = f^{MV\_dir}_{t+1}(d)(X))$ based on specificity level of $p_1$; $p_2$ does not rely on actual values to which the feature evaluates.
In such a case the system prioritizes $p_2$ in the confirmation queue, if $p_1$ gets enqueued at all with confidence threshold. 

For linguistic translation, we use a mapping from qualitative features and time interval of patterns to linguistic descriptions that, in turn, are composed to complete natural language sentences using the rule-based slot-filling mechanism in the interactive interface \citep{krishnaswamy2016voxsim} (see below).

\subsection{Performing actions in novel situations}

In previous work \citep{do2018thesis}, we have addressed a few different approaches to performing learned actions in a novel situations, e.g., \gls{rl} and search algorithms. We must incorporate the learned pattern consistencies from the previous discussion as constraints in the search space of the execution planning. The algorithm we use for action reenactment is a search algorithm, in which an execution is a chain of planned simple steps, such as ``Move(A,\textit{coordinate})'' or ``Rotate(A,\textit{angle})''. A simple search algorithm will generate a set of random candidate steps on the search space, then qualify whether a new step satisfies the constraints. At the same time, the system can verify if a state satisfies a termination condition and decide to announce the completion of the action. \textit{Best-first} or \textit{beam} searches can both be used in this scenario.

\section{Experiments}
\label{ssec:interface}

\vspace{-1mm}
\begin{figure}[h!]
\centering
\includegraphics[width=1in]{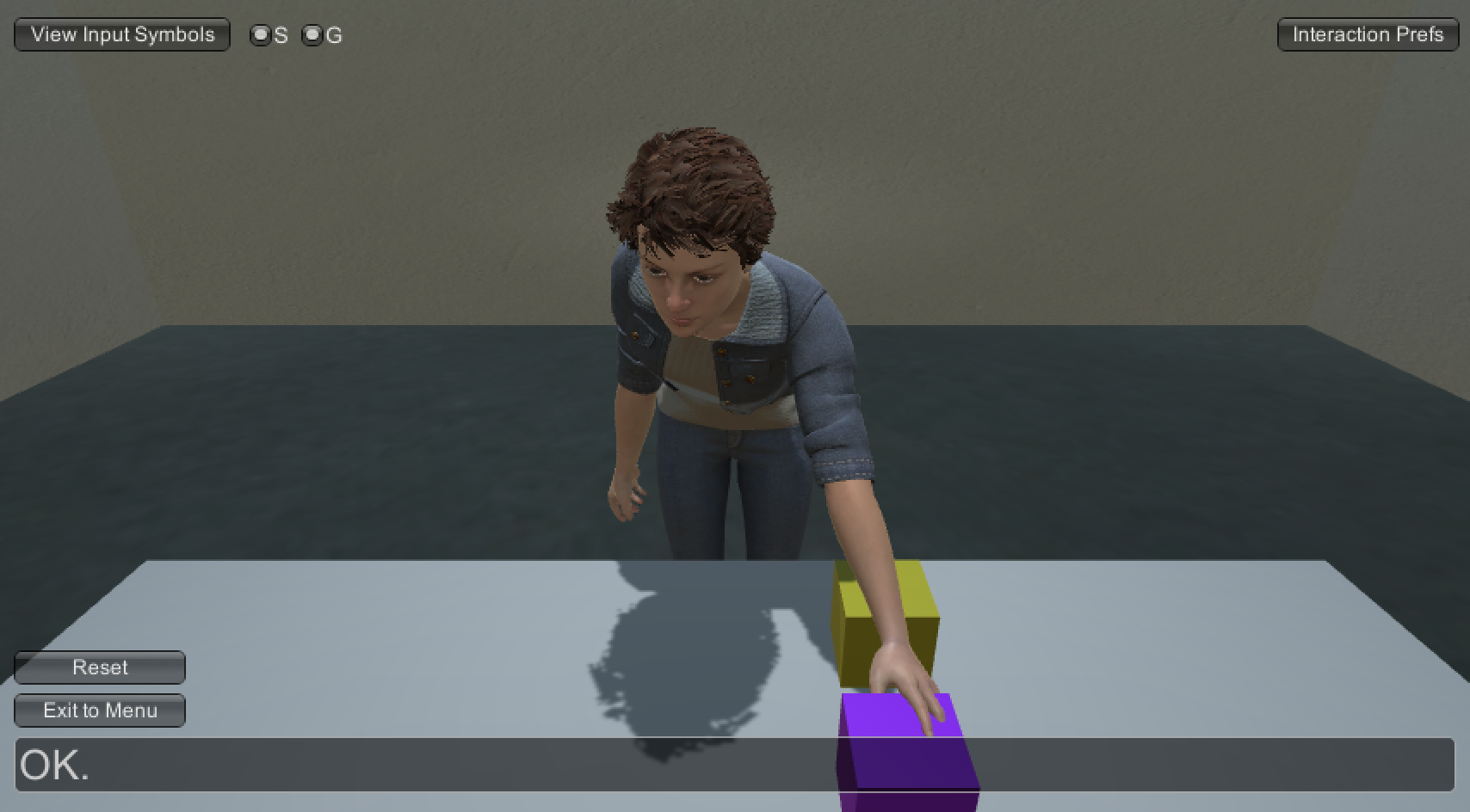}
\includegraphics[width=1in]{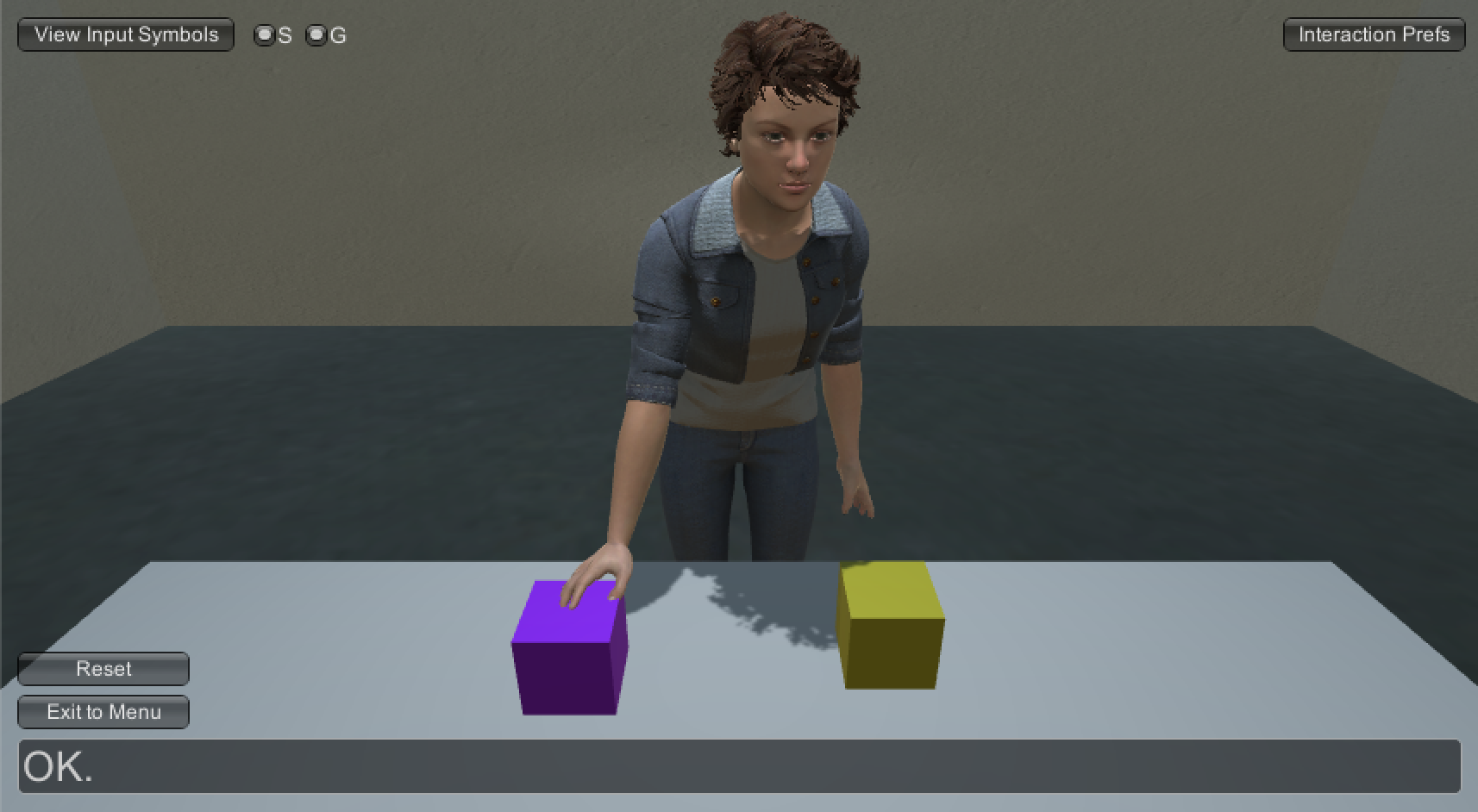}
\includegraphics[width=1in]{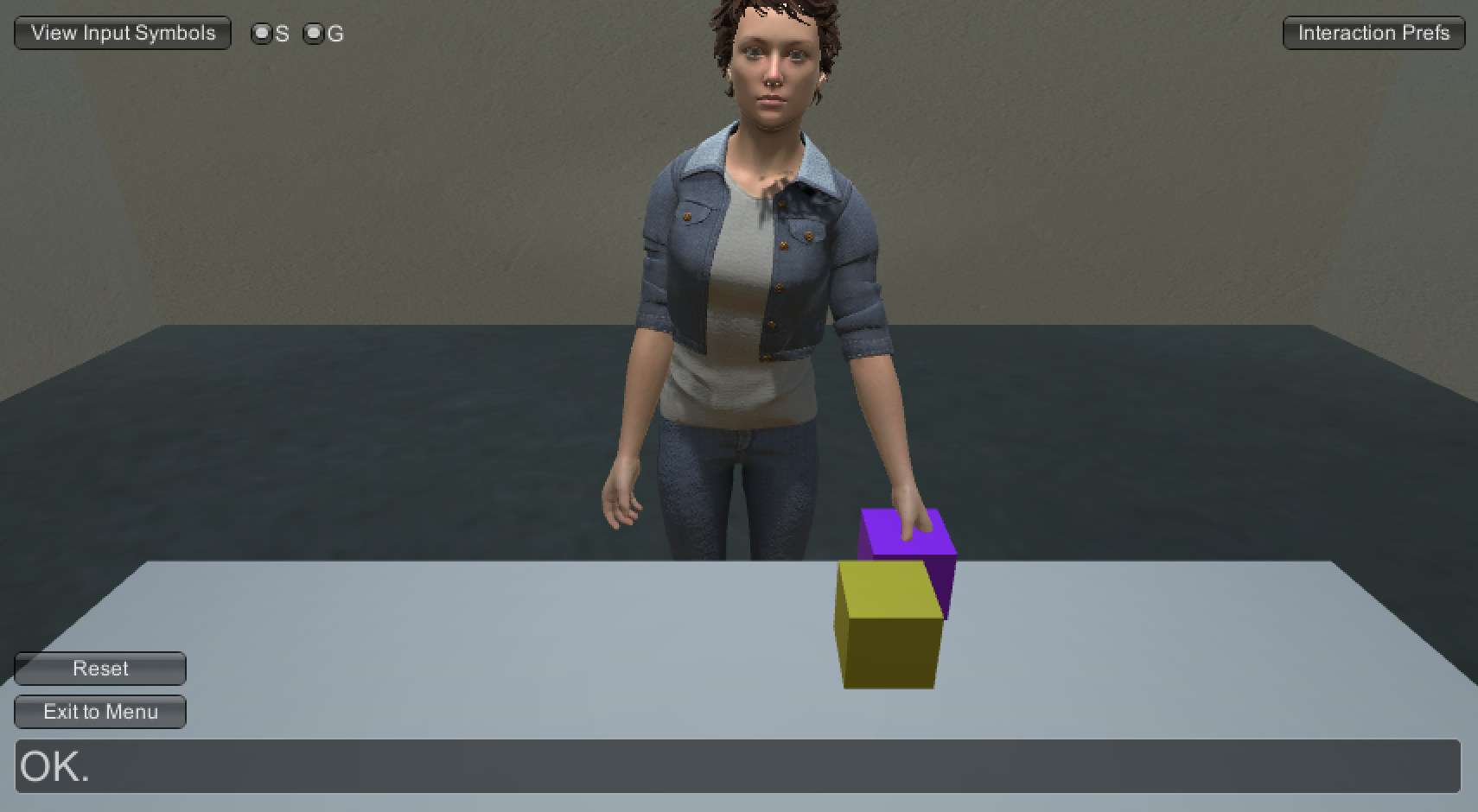}
\vspace{-3mm}
\caption{Sample interaction with avatar}
\label{fig:sample}
\end{figure}
\vspace{-3mm}


Our interaction framework is built on the VoxML/VoxSim platform \citep{pustejovsky2016LREC,krishnaswamy2016voxsim}, which facilitates the encoding of common-sense knowledge about events and objects, and their visualization in a 3D environment as a demonstration of how the computer interprets them.  The interaction system is presented in full by \cite{krishnaswamy2017communicating}.  A human user, standing before a Kinect\textregistered\- camera and a monitor that displays an avatar and a table with blocks on it can, through the use of language and gesture, direct the avatar to perform a set of actions with the blocks.  For instance, by indicating a block (e.g., by saying ``the purple block'' or pointing to the purple block), and then pointing to a new location, the user can direct the avatar to move that block to that new location.

At points in the interaction, the avatar may express uncertainty about an action or need to confirm it.  This may be asking which block the user is indicating, or confirming that the indicated location is the intended destination of the block.  When the avatar encounters a symbol sequence such as $point\_at(block7);point\_at(L1)$ with the existing instruction that the block is to be slid, it can insert these symbols into a predicate structure, e.g., $slide(block7,L1)$ and then disambiguate any piece that it needs to by translating those symbols into a natural language output.  On this primitive level, these simple symbols are encoded in the model, and so with a learning-based approach, the question becomes one of being able to extract more complex symbols in need of confirmation or disambiguation from a learning algorithm, determining what to ask about, and how to phrase the question to prompt a ``yes'' or ``no'' answer.

Each time the avatar completes an action (i.e., finishes moving an object), we write out the complete state of the scene, with the positions and rotations of all blocks.  From this raw data, we extract qualitative spatial relations to be fed into the learning module, using QSRLib \citep{gatsoulis2016qsrlib}.

Some examples of patterns we expect to mine from action demonstrations are given here:

\textbf{Move A around B}

\small{\begin{itemize}
\item $\forall t f^{MV}_{t}(A) \: = \: 1$ and $\forall t f^{MV}_{t}(B) \: = \: 0$
\item $\forall t f^{QTC_{C3}}_{t}(A,B) \: = \: f^{QTC_{C3}}_{t+1}(A,B)$
\item $f^{CD}_{0}(A,B) \: = \: f^{CD}_{F}(A,B)$
\end{itemize}}

\textbf{Make a row of blocks}: we assume that a ``row of blocks" means blocks that are evenly spaced along a single axis. Let us assume that all blocks make a set $\mathcal{S}$. Let us define some functions on $\mathcal{S}$ (that we called dynamic binding in the previous section): $\mathcal{L}(\mathcal{S})$ is the last moved block in $\mathcal{S}$; $\mathcal{C}(\mathcal{S})$ is a selected block for the next move; then we have the follows:

\small{\begin{itemize}
\item $\forall t f^{QDC}_{t}(\mathcal{L}(\mathcal{S}), \mathcal{C}(\mathcal{S})) \: = \: f^{QDC}_{t+1}(\mathcal{L}(\mathcal{S}), \mathcal{C}(\mathcal{S}))$
\item $\forall t f^{CD}_{t}(\mathcal{L}(\mathcal{S}), \mathcal{C}(\mathcal{S})) \: = \: f^{CD}_{t+1}(\mathcal{L}(\mathcal{S}), \mathcal{C}(\mathcal{S}))$
\end{itemize}}

\section{Conclusion and Discussion}

Currently, we are working with data captured from the interactive interface, in which a demonstration has already been broken down into multiple steps (Figure~\ref{fig:sample}). To extend this framework to work with real demonstrations, where data comes in from a continuous stream, we can capture data from real human performances using tools such as ECAT \citep{do2016ECAT}. We may require processing to facilitate pattern mining, including noise removal or ``key'' frame detection (cf. \cite{asfour2008imitation}).

It is also important that the system be able to recognize the relationship between different action concepts. Taking two action concepts, (1) ``Move A around B'' and (2) ``Move A around B clockwise'', as examples, we can see the hierarchical relationship between them implies their corresponding conceptual patterns. Therefore, our system needs to be able to update the learned concept of the first action to be a superclass of the concept of the second action.

We have proposed a system for learning primitive actions using an interactive learning interface. By examining a specific learning scenario, we have demonstrated various requirements of this system.  We believe such a framework can improve on the successes of \gls{lfd} methods by incorporating multimodal information through real-time interaction to facilitate online learning from sparse data to improve models.

\bibliographystyle{aaai}
\bibliography{main}

\end{document}